\documentclass[jair,twoside,11pt,theapa]{article}
\usepackage{jair, theapa, rawfonts}
\usepackage{graphicx}
\usepackage{amsmath}
\usepackage{bm}
\usepackage{caption}
\usepackage{subcaption}

\usepackage{amsmath, amsfonts, algpseudocode, algorithm}
\usepackage[all]{xy}
\usepackage{url}

\newcommand{\eat}[1]{}

\ShortHeadings{Artificial Explanations in Automated Decision-Making}
{Besold and Uckelman}
\firstpageno{1}

\begin{document}

\title{The What, the Why, and the How of Artificial Explanations in Automated Decision-Making} 

\author{\name Tarek R. Besold \email tarek-r.besold@city.ac.uk \\
       \addr Department of Computer Science, City, University of London
			\AND
			\name Sara L. Uckelman \email s.l.uckelman@durham.ac.uk \\
			\addr Department of Philosophy, Durhamn University
}

\maketitle

\begin{abstract}The increasing incorporation of Artificial Intelligence in the form of automated systems into decision-making procedures highlights not only the importance of decision theory for automated systems but also the need for these decision procedures to be explainable to the people involved in them.  Traditional realist accounts of explanation, wherein explanation is a relation that holds (or does not hold) eternally between an \emph{explanans} and an \emph{explanandum}, are not adequate to account for the notion of explanation required for artificial decision procedures.  We offer an alternative account of explanation as used in the context of automated decision-making that makes explanation an \emph{epistemic} phenomenon, and one that is dependent on context.  This account of explanation better accounts for the way that we talk about, and use, explanations and derived concepts, such as `explanatory power', and also allows us to differentiate between reasons or causes on the one hand, which do not need to have an epistemic aspect, and explanations on the other, which do have such an aspect. Against this theoretical backdrop we then review existing approaches to explanation in Artificial Intelligence and Machine Learning, and suggest desiderata which truly explainable decision systems should fulfill. 
\end{abstract}

\section{Explanations and Decisions}
Artificial systems are increasingly used to make decisions in an automatized fashion in various aspects of human life, including medical decision-making and diagnosis, large-scale budgeting, financial transactions, etc.  The range of possible or actual applications of such systems is broad and varied, from the assignment of credits and loans, through recommendations for medical treatments or the distribution of donor organs, to more mundane applications in matchmaking on online dating platforms or the support of healthy or active lifestyles. 
As a result, the important role that theories of decision-making in general, as well as particular implementations via algorithms and decision procedures, play in Artificial Intelligence (AI) cannot be overstated.

But because these artificial decision systems by their nature interact with human users, a robust decision theory or algorithm is not, by itself, going to be adequate.  It is not sufficient that we can merely predict what results some system will obtain, reasoning from first principles of classical decision theory if we do not know why we get those results or are not able to explain how the results are obtained.  There is an important post-decision epistemic gap that must also be filled, when the decision rendered by the artificial system is communicated to ``the human system'': The explanatory gap.

When someone asks for an explanation of why they have been denied a loan by the bank, responding ``the algorithm outputted a `no' to your request'' may be a \emph{reason} why the loan application was denied, but this is often not, and in many cases, \emph{cannot be} an \emph{explanation} of why the person was denied; it isn't an explanation any more than ``Because I said so'' is an explanation to a child of why they cannot have a second piece of candy: ``Because I said so'' is a \emph{reason} why the child cannot have a second piece of candy, but it is not a reason that gives insight into the mechanisms in play; it is simply an appeal to authority.  As Park et al.\ note, ``Explaining decisions is an integral part of human communication, understanding, and learning'' \cite[p.~1]{ParkHASDR16}; an answer that does not produce understanding of why the answer is correct or provide an insight into how the answer was obtained is not going to satisfy the relevant role that explanations play in human communication. In a sense, explanation is the flip side of decision: The capacity to make deliberate decisions brings along with it the need to be able to adequately explain how and why those decisions are reached.\footnote{Many people are good at constructing post-hoc rationalizations of their decisions (especially, but clearly not exclusively, in the case of intuitive, spontaneous and/or ``unconscious'' decisions), but generally we do not count these rationalizations as adequate explanations.}  Thus, a robust and correct algorithm or decision procedure is never going to be enough for satisfactory AI-human interaction: Given that explaining decisions is integral to human communication, it must also be possible to explain why that algorithm or decision procedure gives the outcome that it does.  

The topic of `explanation' is one that is quite frequently discussed in philosophy, particularly in philosophy of science, where, \emph{inter alia}, inference to the best explanation plays a substantial role and the explanatory power of a scientific theory is often cited as a virtue to be promoted when discriminating between theories \cite{harman,thagard}.  Most of these accounts of explanation are realist in nature, grounding explanation in some factor or feature of the real world.  Such accounts of explanation, however, do not seem to hit the mark for the purposes of shedding light on the concept of explanation as it is used in and regarding artificial systems.  We address this in \S\ref{explain}, introducing prominent accounts of explanation found in philosophy and explaining (hah!) why they are inadequate for our current purpose.  In \S\ref{epistem} we make explicit the epistemic dimension of explanations which \emph{must} be addressed in order to have a satisfactory account.  We switch gears in \S\ref{AI} to lay out some relevant AI contexts where an account of explanation is necessary, paving the way for us to highlight desiderata for an account of explanation in AI in \S\ref{desiderata}, building upon the account of explanation we gave in \S\ref{epistem}. We conclude in \S\ref{conc}.

\bigskip

Before we begin, however, we first take a brief look at what it is that we wish to explain, that is, the decisions produced by artificial systems, generally in communication with or relevant to humans, and the ways in which decision theory is manifest in Artificial Intelligence.

Classical Decision Theory (CDT)---``the analysis of the behavior of an individual facing nonstrategic uncertainty'' \cite[p.~2]{gintis}---is rooted in classical game theory and operates under many simplifying assumptions such as transitivity of preferences, unbounded processing capacity, perfect knowledge, and perfect recall.  These assumptions allow decision theorists to construct elegant mathematical models, but these models are poor models of actual human reasoning.  Individual experiences are sufficient to show that humans do not have unbounded processing capacities---an observation which is by no means new, but already lies at the heart of, among others, Simon's work on bounded rationality \cite{simon1959,simon1990}---nor do we have perfect knowledge or recall, and experimental evidence also demonstrates the many ways in which humans fail to be perfect \cite{WC}. This provides fair reason to reject CDT as a plausible (or even ``just'' reasonable) model of human behavior.

It might be thought that CDT fares better in the artificial domain, for many of the simplifying assumptions that do not apply to humans may apply to artificial systems. However, while CDT might offer a convenient framework guiding the development of decision capacities in AI systems which are operating by themselves, the situation changes once systems are required to interact with human users in a significant way. In these cases, the capacity to not only reason rationally, but rather to emulate human-like reasoning becomes important \cite{besold2018}. This does not only hold in scenarios where close human-machine collaboration in a bidirectional way is required \cite{besold2013,besold2018}, but also in the setup serving as backdrop to this article, i.e., when human users are subject to automated decision-making. While in the latter case one can argue that from a purely rational perspective taking decisions based on a CDT model is advisable (given the well-understood mathematical basis of CDT accounts of decision-making, and the corresponding adherence with explicit rationality postulates), once one also takes into account the perspective of the subject of the decision---and their need to not only rationally understand potentially negative decision outcomes, but also to accept them on a personal and subjective level---additional requirements tying more closely into cognitive, psychological, and cultural aspects of decision-making come into play.  

Even models that build on the application of classical game theory to the evaluation of actual human behavior and reasoning --- such as the \emph{beliefs, preferences and constraints} or \emph{BPC} model of behavioral game theory --- weaken many of the assumptions made by classical game theorists, but still adopt problematic axioms. For example, while the BPC model does not require perfect knowledge or unbounded rationality, it still assumes that human preferences are consistent, an assumption that human-based disciplines such as psychology generally reject \cite[p.~3]{gintis}.

These and other shortcomings of CDT---when thinking about extending its reach beyond a descriptive-explanatory use in understanding economic phenomena, or in supporting normative economics---have by now been widely recognized (see, for instance, \cite{beach2017,einhorn1981,gilboa}). Still, it is an open question how CDT would have to be expanded to be usable, for instance, to applications (possibly even predictively) modeling actual human decision-making on a case by case basis. Gilboa notes that
\begin{quote}one may find that refinements of the theory depend on specific applications. For example, a more general theory, involving more parameters, may be beneficial if the theory is used for theoretical applications in economics. It may be an extreme disadvantage if these parameters should be estimated for aiding a patient in making a medical decision. Similarly, axioms might be plausible for individual decision-making but not for group decisions, and a theory may be a reasonable approximation for routine choices but a daunting challenge for deliberative processes \cite[p. 2]{gilboa}.\end{quote}
Modern decision theory has correspondingly developed into a wide and varied field of partially independent, partially rivaling paradigms and methods with usually fairly well-specified application domains and contexts, also in this sense having departed from many CDT approaches and their corresponding claims to generality. 

Examples of this new brand of decision-theoretic frameworks include Gilboa and Schmeidler's subject-centric accounts \cite{gilboa_schmeidler_2001}, in which the rationality or irrationality of a decision depends on the decision-maker's individual cognitive capacities and limitations, or their theory of case-based decision-making \cite{gilboa1995} which leaves aside beliefs or predictions and judges the desirability of an act exclusively based on how well it worked on similar problems in the past. Tversky and Kahneman \cite{tversky1974judgment} in their heuristics and biases program aimed to explain and conceptually rebrand what had been considered irrational human behavior as simply adhering to certain patterns and shortcuts common to human reasoning. Following in a similar vein, behavioral decision theory \cite{slovic1977} and, more recently, behavioral economics \cite{camerer2011behavioral} enhances the repertoire of economics with a wide range of methods and theories from experimental psychology, neuroscience, cognitive science, and the social sciences in providing explanations across different conceptual levels from neurophysiological to cultural for previously unconsidered or dismissed particularities of human decision behavior.

In all its diversity, modern decision theory acknowledges the need to take into account the decision-maker who, often, also turns out to be one of the decision subjects. In this way, modern decision theory shares a basic intuition also driving research into explanations in decision-support and automated decision-making systems. The decision-maker (in the case of decision-support systems) and/or the decision subject require understanding of the why and how a decision has been reached, or a recommendation regarding a decision outcome has been given. This understanding can either be provided if the decision has been reached by modes of reasoning familiar to the user (i.e., the initially mentioned emulation of human reasoning), or by providing explanations of the reasons and the steps which have given rise to the system output in a way understandable to the user. While, all efforts in decision theory (modern or otherwise) notwithstanding, the former option still remains largely out of reach, the latter angle is taken by most current projects in the context of explainable AI systems for decision-making. But despite the enormous popularity the corresponding research direction currently enjoys, some fundamental questions hitherto remain unanswered (and, often, even mostly unaddressed)---such as, for example, what is or makes for an explanation of automated decision-making (and maybe even a good one) in the first place?


\section{What are explanations (thought to be)?}\label{explain}
Standard philosophical accounts of explanation are robustly realist, taking `explanation' to be some metaphysical characteristic of the world.  A typical example of this approach is taken by Nozick, in a chapter entitled ``Why is there something rather than nothing?'' \cite{nozick}.  There he notes that ``the question about whether everything is \emph{explainable} is a different one'' \cite[p.~116]{nozick} (emphasis added) from the title question of the chapter.  Nozick stipulates the existence of a relation $E$ which is the relation ``\emph{correctly explains} or \emph{is the (or a) correct explanation of}'' \cite[p.~116]{nozick}, and states that this relation is irreflexive, asymmetrical, and transitive:
\begin{quote}Nothing explains itself; there is no $X$ and $Y$ such that $X$ explains $Y$ and $Y$ explains $X$; and for all $X$, $Y$, $Z$, if $X$ explains $Y$ and $Y$ explains $Z$, then $X$ explains $Z$ \cite[pp.~116--117]{nozick}.\end{quote}
As a result, this relation of explanation strictly partially orders all truths.\footnote{A strict partial order is an order that is irreflexive, asymmetric, and transitive.}  Further, this ordering is of \emph{all} explanations, not just those that are known to us \cite[p.~117]{nozick}. Such an account of explanation is realist in the sense that 
\begin{quote}if one hopes to explain the occurrence of one event $e$ by appealing to another event $c$, the explanation is successful only if there is a genuine relation $R$ between the mentioned events.  That is, for such an explanation to be \emph{correct} and therefore genuinely explanatory it must actually be the case that $c$ and $e$ stand in relation $R$ \cite[p.~75]{campbell08}.\end{quote}
Unfortunately, this explanation of explanation is almost entirely useless (or at least inapplicable) as an analysis of \textit{practical} explanations. With this we mean the ordinary phenomena of explanation that we believe are relevant for use in the context of automated decision-making in particular, or even when looking at the notion(s) of explanation as used by people in their everyday life more generally. We will highlight a number of issues with Nozick's stipulations, and offer an alternative account.

First, note that Nozick gives no motivation or argument for his account of explanation, either that it is a relation or that it is a relation of the type that he specifies.  He simply asserts that such a relation exists and that it is what constitutes explanation.\footnote{Those who desire an account of why explanation is a relation can look to \cite{woodward}.}

Second, his definition of explanation is incomplete because it does not specify what the relata of the relation are---facts?  States of affairs? Objects in the world? Propositions? Something else?  It may be the case that that it doesn't matter, and his account works whatever the relata are, but we have not yet been given any indication that this is the case.   At the very least, Nozick needs to specify what he thinks the relata are, and, even better, to give us reason to believe that he is right.\footnote{Others have attempted to answer this question. For the ``events and relations \emph{in the world}'', rather than ``items in our epistemic corpus'' \cite[p.~208]{campbell} answer to the ``between what?'' question, see \cite{kim}.}  

Third, Nozick gives no argument for his claim that the relation of ex\-plan\-a\-tion---even assuming that one exists and that the relata of the relation are well-specified---is irreflexive, asymmetrical, and transitive.  In fact, there are reasons to think that it is none of these.\footnote{It might be objected here that the following examples beg the question against Nozick's argument.  The problem is that \emph{Nozick gives no argument}, so it is impossible for us to beg the question against his (non-existent) argument.  He presents these features of the relation of explanation as if they are obvious; the putative counterexamples we raise should at least cast some doubt on this.} If there are brute facts (a possibility that Nozick himself entertains \cite[pp.~117--118]{nozick}), then these brute facts are their own explanations, hence the relation of explanation would be reflexive for brute facts.  In the case of equivalent statements, there is no reason to think that they could not be explanations of each other; for example, the Axiom of Choice and the proof to Zorn's Lemma can be thought to \emph{explain} Zorn's Lemma, \emph{and} conversely Zorn's Lemma and the proof to the Axiom of Choice can be thought of as an explanation of the Axiom of Choice. Which one is taken to explain the other will depend, in part, on which one the person in question was introduced to first. This highlights one important aspect of explanations that we will discuss more in the next section: Explanations are context sensitive.  Lastly, as with any transitive relation, it would not be difficult to construct a sorites wherein $X_0$ explains $X_1$, $X_1$ explains $X_2$, and $X_i$ explains $X_{i+1}$ for every $i$ up to some $n$, but by the time one gets to $X_n$, $X_0$ is not an explanation of $X_n$.  This again highlights the importance of context sensitivity.

Fourth, this account of explanation does not provide any room to distinguish explanations from reasons.  There are many reasons that can be given why a certain thing is the case; which of these turns out to be an explanation will depend on context, we argue below.

What we have seen from the preceding is that Nozick's account of explanation entirely overlooks what we can call the epistemic dimension of explanation.  Now, this is not to deny Nozick's implicit point that some explanations are known to us and some are not---in fact, quite the opposite.  For it is precisely the fact that there are explanations that we do not have that we wish to have that we ask the question ``Why?''  But this question is never asked in isolation, and \emph{that} is the epistemic dimension we are interested in.

\section{The epistemic dimension of explanation}\label{epistem}
Suppose someone asks you ``Why is that car red?'' 
There are a number of possible answers you might give:
\begin{enumerate}
\item Because it reflects light in wavelengths between approximately 625--740 nm.
\item Because someone painted it red. 
\item Because no one painted it blue.
\item Because its owner's favorite color is red.
\item Because there were no non-red cars at the dealer when the owner bought their car.
\end{enumerate}
All of these can plausibly count as reasons for why the car is red. But which of these reasons will count as an \emph{explanation} of why the car is red will---as exemplified by the examples above---depend on the circumstances in which the asker is asking the question.\footnote{For the ``Why is that car red?'' example above, these could, for instance, be: \begin{enumerate} \item In the context of a physics class in school, watching cars drive by on the street. \item In the context of a police search for a blue car of a certain type with a given number plate, after finding the targeted car which turns out to be red. \item In the context of a repainting effort in an autoshop, processing an order to convert red cars into blue ones. \item In the context of a car dealership, selling a red version of a usually blue production model. \item In the context of a conversation about the new car of a person who usually prefers other colors than red.\end{enumerate}}  These circumstances include the asking agent's belief state and knowledge set, as well as her reason for asking that question, as opposed to another question, and what she hopes to do with the answer once she's obtained it.

These factors are what make up what we call the `epistemic dimension' of explanation.  We are not the first to highlight the importance of this dimension; Kim \cite{kim} argues that many ``existing accounts of explanation\dots neglect the epistemological dimension of explanation by failing to provide an account of understanding'' \cite[p.~213]{campbell}.  And as McLaughlin says, ``Only by taking into account the epistemic dimension of explanation can we capture the idea that explanations provide understanding, answer questions, and give reasons for belief'' \cite[p.~227]{mclaughlin}.  While explanations provide reasons, as noted above not every reason counts as an explanation, and thus there must be something more to a reason that makes it an explanation.  What that something more is we now attempt to identify.

First, note that when we talk about the ``epistemic dimension of explanation'', we do \emph{not} mean the sort of thing that Campbell is talking about here:
\begin{quote}Kim's own understanding of the epistemological dimension of explanation does not actually concern understanding \emph{per se}, or how it is that an explanation generates or contributes to understanding; instead it is a question about what kinds of facts constitute explanatory knowledge \cite[p.~214]{campbell}.\end{quote}
The question of how an explanation generates understanding is a question of epistemology, not a question of explanation.  Similarly, the notion of ``explanatory knowledge'' is narrower than the notion of explanation we are interested in; knowledge implies truth but, as we argue below, practical explanations need not be truthful in order to count as explanatory.

We argue that there are three things necessary for a particular reason to count as a practical explanation in a given context: (1) The reason must be relevant to the purpose of the question; (2) the reason must provide the hearer with the power to act in a more informed way; (3) the reason need not be true, though it does need to be at least an approximation of the truth.  We treat each of these characteristics in turn.

Ad (1): Irrelevant reasons cannot be explanatory.  If you are asking me for an explanation, then you have a particular epistemic need to be filled or epistemic longing to be satisfied. This need circumscribes the possible acceptable answers.  Any answer which does not (attempt to) satisfy this need will not be relevant and cannot be an explanation. Further, not only does the epistemic context of the explanation determine (in part) what reasons are actually explanation, but the type of the explanation matters too, whether it is a formal explanation, a mechanistic one, a teleological one, etc.  As Vasilyeva et al.\ point out, ``Research increasingly supports the idea that (many) representations and judgments are sensitive to contextual factors, including an individual's goals and the task at hand\dots This raises the possibility that judgments concerning the quality of explanations are similarly flexible'' \cite[p.~1]{vasilyeva2017}.  Further, empirical evidence demonstrates that ``the acceptability of teleological explanations relates to conceptual domains, causal beliefs, and general constraints on explanation'' \cite[p.~168]{LC}, and there is little reason to doubt that the same is true for other types of explanation.  All of this comes together to demonstrate that an account of explanation that does not take into account the ways in which context determines the relevance of an explanation will fail to be an account of how explanations actually function.

Ad (2): The reason the type of explanation matters for determining relevance is because the type of explanation affects what you can do with the explanation afterwards.  This brings us to the second characteristic of explanations, and the question of what makes a reason relevant in a given context?  That is to say, what makes some particular reason satisfy an epistemic need, but not another?  The answer is rooted in the notion of ``explanatory power''---when we say, e.g., that one scientific theory has ``more explanatory power'' than another, we are saying that it gives us (or, at the very least, the scientists involved in applying the theory) the power to \emph{do more things}.  We can explain other phenomena, we can make new predictions, we can understand more than we understood before.  Thus, the power of an explanation is rooted in its capacity to allow us to act in a way that we could not have acted otherwise.  Irrelevant reasons do not give us such a power.  If you tell someone who asks you ``Why is that car red?'' that it is because it reflects light of a particular wavelength, but the person who asked for the explanation has no concept of wavelengths or reflection, or indeed of light as an abstract concept, this answer will not be explanatory because it does not allow her to act in a more informed way; the answer is, quite simply, irrelevant because it is uninformative, and it is uninformative because it does not fit within the epistemic context of the asker.  This not to say that such an answer \emph{cannot} be explanatory; it can, in another context.  From this, it is clear that whether something is an explanation varies according to context.

Ad (3): This is perhaps the biggest way in which our account of explanation differs from realist accounts.  On a realist account, the relation of explanation either holds or doesn't hold at all times and only holds (presumably) when there is in fact a genuine connection between the two events.  If the car owner's favorite color is green, then saying that the car is red because red is his favorite color is not an explanation because it is false.

However, truth is an enormously high bar to put on explanation, and in fact explanations that are later determined not to be true can still have enormous explanatory power (in the sense of explanatory power that we defined above).  In our pursuit of scientific progress, ``we do not necessarily replace wrong theories with right ones, but rather look for greater explanatory power'' \cite[p.~93]{niaz}. A classic example of this from philosophy of science are scientific models of the structure of atoms.  Over the course of the 20th century, different models of atomic structure were proposed which ``continue to provide increasing explanatory power, such as: Thomson, Rutherford, Bohr, Bohr-Sommerfeld, and wave-mechanical, among others'' \cite[p.~71]{RepNat}.  It simply doesn't make sense to speak of increasing explanatory power if you think that there is no explanation going on (which one must think if one requires all explanations to be strictly truthful). Here it is worth noting that saying that Theory 2 is adopted to replace Theory 1 because Theory 2 has more explanatory power than Theory 1 does not commit us to saying that Theory 1 retains any explanatory power once Theory 2 has taken its place; in fact, our approach to explanation allows us to strip Theory 1 of all of its explanatory power once its been superceded (though of course we are not required to: Newtonian mechanics are still explanatory, even if in some contexts relativistic mechanics are \emph{more} explanatory).

We do, however, want to encourage truth-seeking in our quest for explanations, and to prioritize as good explanations those which have a better fit with the set of knowledge claims relevant for the context (and are thus in at least some sense ``closer to the truth'').  Consider folk-explanations for thunder, whether it be Zeus throwing his thunderbolt, Thor banging his hammer on an anvil, or Leigong hitting his drum with his mallet.  In the right context, each of these can be relevant to satisfying the hearer's epistemic longing; in such contexts they also provide the hearer with the power to act in a more informed way (for example, one can then consider whether to sacrifice a virgin to appease the god and make the thunder stop); however, as an approximation of the truth each of these explanations all falls quite short.  As a result, it is legitimate for us to say that they are \emph{not very good} explanations.  Thus our account of explanation avoids one potential criticism of non-realist accounts, namely that anything whatever can count as an explanation, given the right context.  Even if that is the case, we are still able to distinguish good explanations from bad ones. 

The question of how much truth is required is a question of great importance in its own right, and one that we will set aside for the remainder of this paper.

\bigskip

We've noted one consequence of this account of explanation above, namely that one and the same reason can be an explanation in one context and not in another, because of the individual nature of individual epistemology.  A further consequence is that one need not give up Nozick's account of explanation as a metaphysical relation between things entirely, of course.  As Campbell notes,
\begin{quote}The pluralist's emphasis on the epistemology of explanation does not render her position irrealist because the correctness of the explanation of one event in terms of another is in part a function of the metaphysical relations between them \cite[p.~91]{campbell08}.\end{quote}
Note, though, that the idea of ``the correctness'' of an explanation potentially smuggles in some problematic notions---not stemming from the ``correctness'' but from ``the''.  Even the pluralist's approach to explanations requires that there be \emph{the} correct explanation, and that this unique explanation's correctness is rooted in certain metaphysical facts (see quote from Campbell earlier in the previous section).  If, however, we scrap the notion of there being \emph{the} correctness of an explanation, and allow there to be many different ways of grounding what makes an explanation a good explanation in a given epistemic context, then we can allow that the existence of some metaphysical relation between events can be sufficient for possessing an explanation, but it is not necessary (and it is not even \emph{always} sufficient).  Thus, we allow for the possibility that we can have multiple possible explanations for a single event or phenomenon, not all of which will be actual explanations in a given context.

In this, our explanatory pluralism differs from Kim's, on which ``it is possible to have more than one explanation for a given event \emph{provided that one has an account of the way the explanations are related}'' \cite[p.~86]{campbell08} (emphasis added).  We can allow that any of the answers to ``Why is that car red?'' above are explanations of the car's being red, without requiring that we have an account of the way in which these various answers are related to each other (and indeed, why would we explain there to be any such account?  Particularly of how people's favorite colors are related to how light at various wavelengths appears to us.)

A final important consequence of this account is that no one can determine whether something is an explanation for someone else, because of the private nature of individual epistemology.  This raises interesting issues in the implementation of mechanisms of explanation into decision procedures in artificial systems, for it means that there is no single answer that the decision procedure can give that can be guaranteed to be explanatory for all people.

Our conclusion is that explanation is ``an epistemological activity'' and explanations are ``an epistemological accomplishment'' \cite[p.~225]{kim88}---they satisfy a sort of epistemic longing, a desire to know something more than we currently know.  Not only do they satisfy this desire to know, they also provide the explanation-seeker a direction of action that they did not previously have.

\section{Explanation in AI}\label{AI}

We now shift our focus to how our conception of practical explanation plays out in the context of AI---both in terms of how it compares to previous accounts of explanation and how well it can play the role needed in AI.  The purpose of this section is primarily historical, outlining what has been said previously as well as the current discussions, before we move on to more normative matters in the next section.

Such a historical discussion is not straightforward: Not only is there no unified or uniform concept of `explanation' that is used in AI contexts, quite often the term is neither defined nor explained.  It is outside the scope of this paper to give a complete history; instead, we focus on two important contexts in which explanations play an important role: explanations in what is called ``Good Old-Fashioned Artificial Intelligence'' or GOFAI (\S\ref{gofai}) and explanations in machine learning (\S\ref{ml}), and then specific challenges concerning explanations that arise in the context of automated decision-making (\S\ref{dm}).

\subsection{Explanation in GOFAI}\label{gofai}

Discussions concerning (the need for) explanations of the reasoning and behavior of AI systems are not a new phenomenon, but already started during the time GOFAI \cite{haugeland1985}, and more precisely in the context of expert and decision support systems. Clancey \cite{clancey1983} questioned whether uniform, weakly-structured sets of if/then associations (i.e., simple inference rules of the form ``IF $precondition_1$ and $precondition_2$ and $\ldots$ and $precondition_n$ THEN $consequence$'') as used in the MYCIN medical expert system for the abductive diagnostics of bacterial infections \cite{shortliffe1974} are suitable for instance in a teaching setup, i.e., with the intent to support active learning. This engendered significant interest in explainable expert systems (see, e.g., \cite{chandrasekaran1989,wick1992}), with much work targeting the representation formalisms used in the respective systems \cite{gaines1996}. It also caused the development of proposals to conceptually split the task that a computational system has to solve into two functional components, a problem-solving one and a communication one (see, e.g., \cite{vansomeren1995,askiragelman1998}).

In a more recent effort originating from the cognitive systems lines of research, Forbus emphasized the importance of the human comprehensibility of the behavior and the output of AI systems in the context of his \emph{software social organisms} \cite{forbus2016}. In his view, participation in human society requires effective and efficient communication.  In order to have both effective and efficient communication, AI systems must have adequate explanation capabilities and capacities.

\subsection{Explanation in Machine Learning}\label{ml}

Machine Learning (ML) methods have seen impressive successes over the last few years. ML-based systems have consequently been introduced into more and more complex application domains, with a significant share of efforts targeting decision support and automated decision-making systems. In the wake of these developments, questions of how to interpret or explain the applied methods and systems have become important. Taking stock of the current variety of ways ``explanations" and related notions are treated within ML as a field, Lipton points out that ``the term interpretability holds no agreed upon meaning, and yet machine learning conferences frequently publish papers which wield the term in a quasi-mathematical way'' \cite[p.7]{lipton2016}. He calls for further formulations of problems and their definitions, hoping to provide a more systematic conceptual basis on which to advance research and development of the corresponding types of systems. Doran et al.\ \cite{doran2017} responded to that call with an initial proposal for a general typology of explainable ML and AI methods and systems. In their account, there are three general types of AI/ML systems:
\begin{itemize}
\item Opaque systems where the mechanisms mapping inputs to outputs are invisible to the user. This basically converts the system into an oracle making predictions over an input, without indicating how and why predictions are made. 
\item Interpretable systems where a user can not only see, but also study and (given potentially required expertise, resources, or tools) understand how inputs are mathematically mapped to outputs.
\item Comprehensible systems which emit symbols along with their outputs, allowing the user to relate properties of the input to the output.\footnote{The definition of the ``comprehensible systems'' category echoes the same intuitions underlying Michie's (much older) notions of {\em strong} and {\em ultra-strong machine learning} \cite{michie88}, cf.\ \S\ref{desiderata}.}\end{itemize}
When comparing the notions of interpretable and comprehensible systems, it is important to note that while interpretable systems are pushing towards becoming ``white boxes'' (in contrast with the ``black box'' nature of opaque systems), a comprehensible system can well remain a ``black box'' concerning its inner workings, but is required to provide the user with symbolic output suitable to serve as basis for subsequent reasoning and action (possibly resulting in a ``communicating black box''.  Against the backdrop of these three types of AI/ML systems, Doran et al.\ require that any definition or characterization of `explanation' must involve the presence of ``a line of \emph{reasoning} that explains the decision-making process of a model \emph{using human-understandable features of the input data}'' \cite[p.7]{doran2017}.  We now see how this plays out in existing ML approaches.

Argument-Based Machine Learning (ABML) \cite{mozina2007} applies methods from argumentation in combination with a rule-learning approach. Explanations provided by domain experts concerning positive or negative arguments are included in the learning data and serve to enrich selected examples. Still, although ABML adds the corresponding information to the system output (and, in doing so, likely enhances the general degree of explanation) compared to most ``standard'' ML approaches, there is no built-in check or guarantee that users fully comprehend the learned hypotheses.

Explanation-Based Learning (EBL) (e.g., \cite{ebg:mitchell}) uses background knowledge in a mainly deductive inference mechanism to ``explain'' how each training example is an instance of the target concept. The deductive proof of an example---in some cases augmented by newly added features not explicit in the training examples \cite{prolog:ebg}---yields a specialization of the given domain theory leading to the generation of a special-purpose sub-theory described in a user-defined operational language. Even so, the generated syntactic explanations can still be far from human-comprehensible explanations in any relevant semantic sense (causal, mechanistic, etc.).

In the context of artificial neural networks and related statistical approaches, regression models \cite{schielzeth2010} or generalized additive models \cite{lou2012} often serve as prime examples for interpretable methods and systems. In these cases, however, interpretability refers almost exclusively to a mathematical property of the models \cite{rudin2014,vellido2012}, allowing for a certain degree of knowledge extraction from the  model and subsequent interpretation by domain experts, but clearly lacking a general explanatory component accessible to the end user.  This is not always problematic; for internal purposes, this notion of interpretability may suffice.  However, once these systems begin to interact with humans, an explanatory component becomes necessary.

Finally, a somewhat popular explanation strategy is to create a more comprehensible representation of the learned model, which in most cases necessitates a trade-off between fidelity and comprehensibility \cite{vandemerckt1995}. Here, examples include the simplification of decision trees via pruning \cite{bohanec1994}, or the extraction (``distilling'') of decision trees \cite{frosst2017}, or $M$-of-$N$ rules\footnote{An $M$-of-$N$ rule is a classification rule of the form ``IF ($M$ of the following $N$ antecedents are true) THEN $\ldots$ ELSE $\ldots$'' \cite{towell1993}. $M$-of-$N$ rules offer a way to succinctly express, for instance, parity problems like the XOR classification problem: ``IF (exactly $1$ of the $2$ inputs is true) THEN odd parity ELSE even parity''. } \cite{odense2017} as more explanatory models from neural networks. What is common to all methods and systems following this route to explanation is that the respective approaches are purely intended to illustrate the system's behavior to the end user while abstracting away from the actual details of the underlying algorithm. This presupposes an application scenario and/or intended user base which afford this restriction on the accuracy of the explanation regarding the actual inner workings of the system, including the potential challenges this might pose, for example, in the context of legal liability and responsibility considerations. Also, while the resulting representations are intended to be more comprehensible than the learned model in its original form, they very often still remain quite technical in appearance and (once again) presuppose a fairly high degree of expertise on the side of the user to actually be understood.

\subsection{Explanation and automated decision-making}\label{dm}

Having looked at AI/ML in terms of technical approaches to explainable or interpretable methods and systems, in this section we focus on the specific conceptual role and challenges regarding explainability arising from the use of AI/ML methods in systems for decision-support and, even more importantly, automated decision-making.  As noted in the introduction, these systems are used across a wide range of applications.  What is common to the vast majority of such systems is their partial or complete reliance on statistical---and, as such, necessarily data-driven---approaches to solving the respective task. This central role of large amounts of data as key input element, processed using complex statistical methods without the explicit generation of interpretable knowledge along the way, gives rise to a certain form of opacity from the user's perspective. They are opaque ``in the sense that if one is a recipient of the output of the algorithm (the classification decision), rarely does one have any concrete sense of how or why a particular classification has been arrived at from inputs. Additionally, the inputs themselves may be entirely unknown or known only partially'' \cite[p.~1]{burrell2016}.  This opacity is in problematic tension with the goal of having explanations for the outcomes of the decision-making system.

This perceived opacity can go back to at least one of three roots \cite{burrell2016}: (1) A system has been designed to be opaque as a consequence of intentional corporate or state secrecy, aiming at self-protection and concealment and, along with it, introducing the possibility for knowing deception. (2) The system is opaque due to technical illiteracy, since reading and writing code---which might be required in an attempt to analyze the system---is a specialist skill. (3) The opacity arises due to the way the respective algorithms operate at application scale, rooting in the mismatch between mathematical optimization in high-dimensionality on the one hand, and the limitations on human-scale reasoning and the corresponding styles of semantic interpretation on the other.

On the one hand, intentional secrecy can be hard to solve; and in some cases, a solution might not even be desired.  On the other hand, opacity due to technical illiteracy and opacity due to processing differences between AI/ML algorithms and human reasoning---while very different in the nature and quality of underlying ailment---can both be addressed by equipping decision systems with explanation capacities. Wachter et al.\ point out that in a systems context, these explanations can operate on one of two levels, either operating on the level of the decision mechanism itself (i.e., targeting the system functionality in terms of the logic, significance, envisaged consequences and general functionality of an automated decision-making system), or on an instance level (i.e., targeting individual decisions in terms of the particular rationale, reasons, and individual circumstances of a specific automated decision) \cite[Sect. 2]{wachter2017}. This distinction between mechanism- and instance-based explanation of an automated decision also becomes relevant when thinking about the timing of the explanation relative to the explained decision-making process. If an \textit{ex ante} explanation is required prior to the automated decision-making process, the resulting explanation necessarily can only address the mechanism level.\footnote{As also noted by Wachter et al.\ \cite{wachter2017}, a small qualification is in place here: If sufficiently simple, pre-defined models are used and are completely specified and known \textit{a priori}, predictions about the rationale of a specific decision become possible in principle already prior to the actual automated decision-making process.} An \textit{ex post} explanation after the decision process has terminated can take both levels into account, potentially addressing aspects of the system functionality as well as the rationale of the specific decision.

Putting these conceptual considerations into the context of actual AI systems as currently (and likely in the short to midterm) deployed for automated decision-making, in the vast majority of cases a divergence between a system's level of explainability and its performance levels has to be noted. At the moment, Deep Learning (DL) approaches \cite{lecun2015} frequently solve tasks which had previously been considered far out of the reach of contemporary AI systems, and raise the bar on most technical benchmarks to which some of the corresponding methods can be applied. Still, this class of ML techniques is almost exclusively made up by methods which even on the level of individual decision instances are hardly interpretable, and which---for reasons tying into the very nature of the corresponding approaches as methods learning their own internal representations\footnote{An important factor in the success of DL approaches to ML is the capacity of the artificial neural networks to learn their own internal representations for relevant input features, making use of the numerous degrees of freedom offered by the high number of network layers. Still, at the moment there is no method assuring that the content of these internal representations refers to anything humans would recognize as meaningful in their conceptualization of the world. Claimed correspondences between the representation layers of deep networks and human conceptualizations are either accidental or misleading.}---generally cannot by themselves provide comprehensibility queues. Thus, while not being completely opaque in that an instance-based interpretation of a decision process might in principle be possible, the actual level of insight into the reasons and the process underlying a decision offered in practice is low. Better interpretable (or even explainable) approaches such as, for instance, the ones discussed in \S\ref{ml} in general fall short of the performance levels exhibited by DL methods, requiring the system's architect to exchange increased interpretability or explainability for (often significant) losses in terms of the system's effectiveness and efficiency, possibly up to a level where a task could not satisfactorily be solved anymore. Concerning potential practical implications of this trade-off, in absence of regulatory requirements or market-relevant incentives to the contrary, providers of AI systems for automated decision-making in a competitive economic environment are likely to prioritize performance over other considerations---producing oracle-like systems, with very high statistical certainty returning ``correct'' (as compared to some externally defined task- and/or domain-specific criteria) decision outcomes but not providing actually informative insight into why and how the decision was reached.

In the context of current initiatives on the side of the regulatory authorities on different levels (such as, e.g., the EU General Data Protection Regulation 2016/679) and societal discussions regarding a desire for transparency and corresponding accountability of automated decision systems, work on better interpretable or explainable methods and systems in AI and ML is ongoing. It is against this backdrop, combined with our considerations concerning the nature and virtues of practical explanations, that we want to have a look at what are desirable properties of explanations provided by AI systems in the following section. The resulting list of desiderata can then help to guide the further development of methods and systems towards the goal of providing actual explanations to the subjects' of automated decision-making.

\section{Desiderata for explanations in AI}\label{desiderata}
In the early days of ML, Michie \cite{michie88} introduced a three-class categorization of learning systems, which has been given new relevance by the current discussion concerning the  explainability of AI/ML systems. In Michie's view, ML systems can be categorized by adherence to the following three (increasingly demanding) criteria:
\begin{itemize}
\item \textit{Weak machine learning}: The system's predictive performance improves with increasing amounts of data.
\item \textit{Strong machine learning}: In addition to meeting the weak criterion, the system provides its learned hypotheses in symbolic form.
\item \textit{Ultra-strong machine learning}: In addition to meeting the strong criterion, the presentation of the hypotheses has to be communicatively effective in that the user is made to understand the hypotheses and their consequences, subsequently improving the joint performance to a level beyond that of a user studying only the training data.
\end{itemize}
Whilst most modern ML systems meet the first, weak, criterion, it is the third (i.e., ultra-strong) demand that resonates most strongly with our discussion of explanation in \S\ref{explain} and \S\ref{epistem}. From a pragmatic point of view, it seems necessary for an explanation to be communicatively effective. In an applied scenario, an explanation only counts as an explanation if it also fulfills an explanatory function.

A similar intuition, though augmented by a second communicative direction back from the user to the system---converting the previously one-dimensional communicative situation into a real two-way interaction also in terms of information transfer, and not only in joint behavior---underlies the account given by Stumpf:
\begin{quote}First, the system's explanations of why it has made a prediction must be usable and useful to the user. Second, the user's explanation of what was wrong (or right) about the system's reasoning must be usable and useful to the system. Both directions of communication must be viable for production/processing by both the system and the user \cite[p.2]{stumpf2009}.
\end{quote}
The demand for effective communication in the direction from the user to the system in our account goes beyond the requirements a system would have to meet to be considered explainable. Still, one of the underlying theoretical requirements for such an exchange to be possible in the first place is worth to be noted: It must be possible for the user to process the explanation provided by the system in such a way as to be able to point out where the system's reasoning in producing the prediction and the subsequent explanation was right or wrong. This goes beyond the mere requirement of the explanation being usable and useful, but poses a stronger demand in terms of content, structure, and presentation of the explanation. 

Several of these aspects also resonate, for instance, in Bohanec and Bratko's observation that in the context of explainable AI, simple, though possibly not perfectly accurate definitions of concepts (demanding for the definition to correspond to the concept in a sufficient rather than a perfect manner) may well be more useful than completely accurate, but complex and very detailed ones \cite{bohanec1994}. Regarding the technical realization of these and similar ideas in an AI system's architecture, Van de Merckt and Decaestecker \cite{vandemerckt1995} suggest conceptualizing systems as two-layered, with a deep knowledge level optimized for the actual task the system is supposed to solve, and a shallow knowledge level optimized for comprehensibility, addressing a description task targeting the deep knowledge level's output. Both levels are connected by an interpretation function from the deep to the shallow knowledge level, allowing one to build an approximately correct but comprehensible description. 

Doran et al.\ \cite{doran2017} discuss external demands which often are brought forward regarding properties for explainable AI systems, including confidence, trust, safety, ethicality, and fairness. Still, as pointed out in their paper, requirements such as to instill confidence and trust that a system's output is accurate are deemed problematic due to their subjective nature,  either depending on users' internal attitudes towards AI systems, previous experiences when using these systems, or on cultural and societal norms and standards. Finally, they also also dismiss completeness of explanations as a required trait---and even question the desirability of complete explanations in many scenarios---pointing to the example of a doctor presenting an incomplete explanation to a patient, either taking into account the patient's limited knowledge of potentially complex biological processes, or sparing her worrisome but ultimately irrelevant details \cite{doran2017}. 

Summarizing the current literature, we find at least two main desiderata regarding explanations in the context of AI and automated decision-making. Each of them constitutes a necessary criterion regarding the status of a system's output as explanation, though none of them is sufficient by itself. First, the explanations a system provides for its reasoning and behavior have to be communicatively effective relative to the system's user, both in content as well as in presentation (``\textit{communicative effectiveness}''). Users must be capable to understand both the presented explanation, as well as its ramifications, in such a way as to be empowered to subsequently adapt their interactions with the system in a beneficial way. Second, the explanations a system provides must be sufficiently accurate (as opposed to perfectly accurate) relative to the explanans and to the context of the system and its user (``\textit{accuracy sufficiency}''). It might be worth trading off some accuracy for improved comprehensibility of the resulting explanation for the user, supporting the communicative efficiency of the explanation. Based on our previous analysis of what makes a practical explanation, we add another two necessary criteria to the list. Third, the explanations a system provides must be sufficiently truthful (as opposed to perfectly truthful) (``\textit{truth sufficiency}''). On the one hand, a trade-off similar to the accuracy vs. comprehensibility consideration might be required or even desirable, while on the other hand considerations akin to the doctor example from the previous paragraph also might warrant to opt for a not perfectly truthful explanation. Fourth, the explanations a system provides must quit the respective user's subjective epistemic longing (``\textit{epistemic satisfaction}''). It is not fully sufficient to meet a user's epistemic needs (which are mostly addressed by the conjunction between the initial two desiderata), but the user also must indeed be under the impression that her search for an explanation has been completed successfully. Taking all four desiderata together, the combination between communicative effectiveness, accuracy sufficiency, truth sufficiency, and epistemic satisfaction of the users for us provides a sufficient characterization of what is needed for an AI system's output to constitute an effective explanation to its users.

\section{Conclusion}\label{conc}
Any discussion of the implementation of decision theory into artificial systems or AI research cannot overlook the importance of the role that explanations play in automated decision-making.  Due to the ``imperfect'' nature of human beings when held to the normative standards set by classical models of decision-making, the latter are inadequate for providing decisions which can be explained in real-life contexts.  A second factor that contributes to this difficulty is that practical explanation is---contra what many more metaphysically-oriented philosophers have argued---best understood not as an abstract relationship that always holds or never holds between two events or facts, but rather has an epistemic dimension that means what counts as an explanation varies by context.  Many things constitute this epistemic dimension, including the knowledge of the person requesting the explanation, the notion of `epistemic longing', and the need for explanations to provide the receiver with the power to act in a way that she would not have otherwise been able to act. Keeping the importance of this epistemic dimension in mind, and looking at previous approaches to constructing explainable AI systems, it turns out that current methods are not sufficient yet. Many efforts have been and are being undertaken to increase the explainability of automated decision systems, with different techniques focusing on different aspects of what constitutes a practical explanation. Still, what is hitherto lacking are clear criteria for explainable AI systems which are conceived in a way so that they can serve as guiding beacons for the corresponding developments in AI theory and engineering. with this article we aim to contribute to closing this gap by putting four candidate desiderata up for discussion: communicative efficiency of the system relative to its users, a sufficient degree of accuracy and a sufficient degree of truthfulness of the provided explanations, and the need to quit a user's epistemic longing.

\section*{Acknowledgments}
We would like to thank Lorijn Zaadnoordijk for her feedback and the many conversations on the topic, and Gwen Uckelman for valuable inspiration.

\vskip 0.2in
\bibliographystyle{theapa}      
\bibliography{explanation}   

\end{document}